\documentclass{arxiv_lumiscrib}

\usepackage{url}
\usepackage{xspace}
\usepackage{wrapfig}
\usepackage{array}
\usepackage{simpleicons}
\usepackage{enumitem}

\AddToHook{cmd/paragraph/before}{\par\vspace{-2mm}}

\newcommand{\method}{\textsc{Dior}\xspace}

\definecolor{oursblue}{RGB}{40,90,158}

\providecommand{\rev}[1]{#1}

\title{\textbf{D}rawing the L\textbf{i}ght \textbf{O}f Image via Material \\
Decoupled Illumination \textbf{R}epresentation\\[2mm]
{\normalsize\normalfont
\href{https://github.com/KlingAIResearch/Dior-Light}{\simpleicon{github}\,\texttt{GitHub}}
\quad
\href{https://huggingface.co/spaces/Little-ECHO/dior-light}{\simpleicon{huggingface}\,\texttt{Hugging Face Demo}}}}
\author[1,2,3,\ddagger]{Xuanpu Zhang}
\author[2,\dagger]{Xuesong Niu}
\author[2]{Haoxiang Cao}
\author[3]{Jianhao Zeng}
\author[3]{Ruidong Chen}
\author[2,*]{Changqian Yu}
\author[1,*]{Shiguang Liu}
\affiliation[1]{Tianjin University}
\affiliation[2]{Kolors Team, KlingAI Research}
\affiliation[3]{Deva Research}
\contribution[*]{Corresponding Authors}
\contribution[\dagger]{Project Leader}
\contribution[\ddagger]{This work was conducted during the author's internship at the Kolors Team.}
\date{arXiv version, October 2026}

\hypersetup{
  pdftitle={Drawing the Light Of Image via Material Decoupled Illumination Representation},
  pdfauthor={Xuanpu Zhang, Xuesong Niu, Haoxiang Cao, Jianhao Zeng, Ruidong Chen, Changqian Yu, Shiguang Liu}
}

\abstract{Controllable image relighting is an important problem in image editing, 
and hand-drawn scribbles provide an intuitive interface for specifying 
the desired illumination. However, existing methods do not establish a 
consistent and effective mapping between scribble inputs and relighting 
results, limiting their ability to control illumination intensity, 
chromaticity, and complex spatial distributions. We address this limitation 
by introducing a material-decoupled illumination representation, 
termed the Lumi Map, which establishes an explicit mapping between 
user scribbles and the resulting illumination, thereby improving 
both relighting accuracy and controllability. Specifically, we use 
a renderer to synthesize source image-Lumi Map-relit image triplets 
and train the model to predict the target relighting result 
conditioned on the Lumi Map. To mitigate the domain gap introduced 
by synthetic data, we further perform reconstruction training on 
real relighting pairs, improving the model's generalization to real-world 
images. Finally, we present \textbf{Dior-Light}, an image relighting method 
controlled by hand-drawn strokes. Extensive experiments demonstrate that 
our method outperforms existing approaches in relighting accuracy and enables effective 
control over illumination intensity and chromaticity on in-the-wild images.
}

\begin{document}

\maketitle

\begin{figure}[!ht]
\centering
\includegraphics[width=0.99\linewidth]{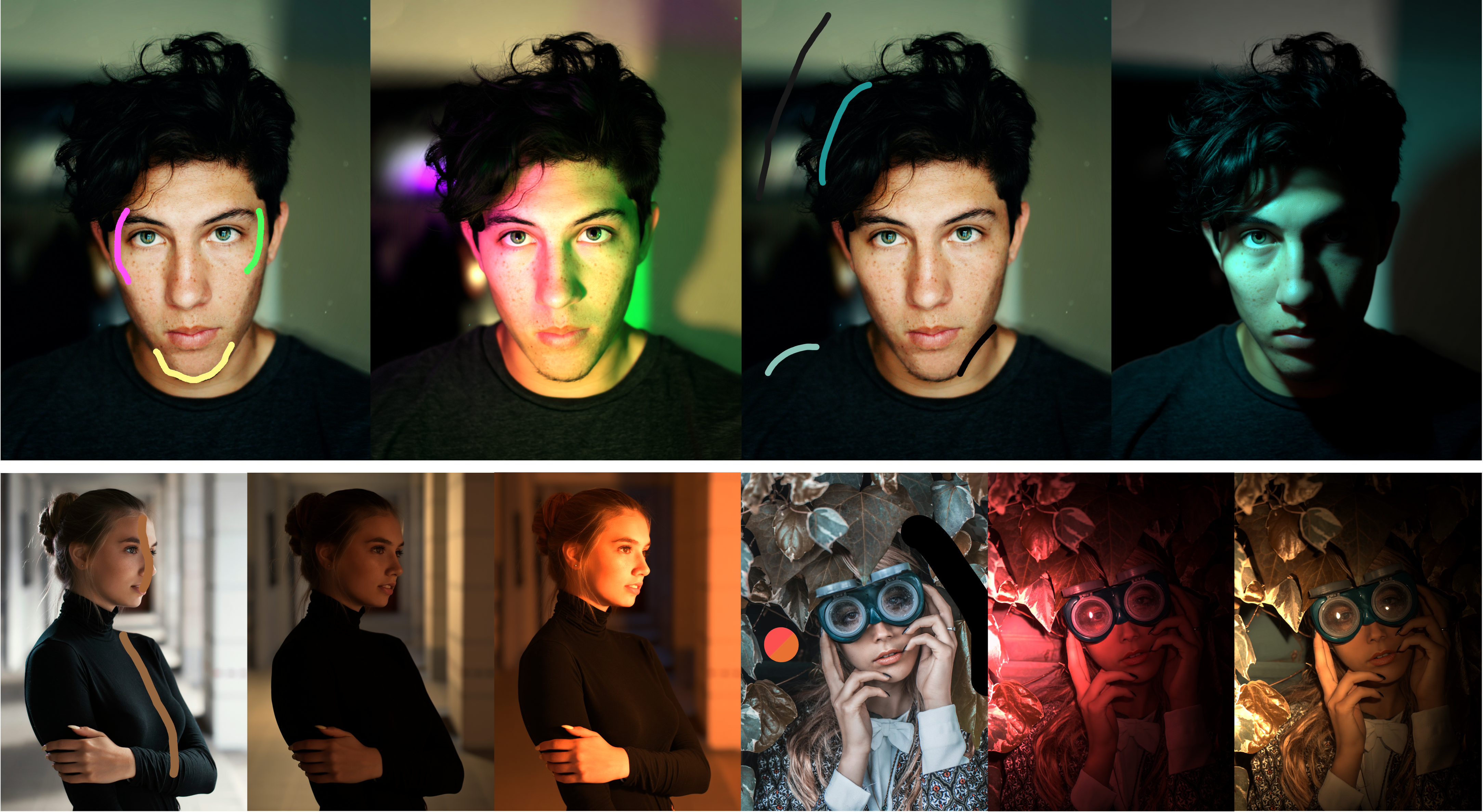}
\caption{\textbf{Controlling illumination in images with hand-drawn strokes.}
Our method enables users to directly specify illumination in an image through simple
strokes. By adjusting the stroke values, users can achieve material-consistent
changes in both illumination intensity and chromaticity.}
\label{fig:teaser}
\end{figure}

\section{Introduction}
\label{sec:intro}

Image relighting is a longstanding problem in image editing and has advanced rapidly
with modern generative models. Given a photograph, the goal is to change its
illumination while preserving the original viewpoint, geometry, surface reflectance,
and subject identity. Recent diffusion-based relighting methods have substantially
improved visual realism~\citep{iclight,neuralgaffer,dilightnet}. However, precise
illumination control remains a key bottleneck. A straightforward approach to explicit control 
is to decompose an image into illumination-related components, such as an environment map or 
shading, and scene attributes, such as geometry and material properties~\citep{rgbx,intrinsic,nerfactor}. 
The illumination components can then be modified independently before the image is recomposed. 
However, this approach relies heavily on the accuracy of image decomposition, while 
high-quality illumination representations are often difficult to obtain for real-world photographs.

Scribble-driven relighting instead investigates how to control target illumination using sparse 
lighting cues, enabling users to interactively adjust the relighting result through hand-drawn scribbles. 
Existing methods have demonstrated that the rich priors of generative models can achieve global relighting 
from local, sparse lighting conditions~\citep{scribblelight,lightpainter}. However, such control is 
primarily expressed through the spatial distribution of illumination, while other lighting attributes, 
such as intensity and chromaticity, remain less explicitly controllable.

Achieving accurate and controllable scribble-driven relighting requires 
addressing two fundamental challenges. (a) Predicting a complete 
illumination distribution from a sparse illumination representation is inherently
underdetermined. Without a well-defined mapping between the lighting scribble and 
the target illumination, precise control cannot be achieved. During inference, this 
ambiguity may manifest as drift in lighting attributes or as relighting results that 
remain constrained by the distribution of the training data. (b) Training data with 
physically consistent light transport are difficult to obtain. Triplets consisting 
of an input image, a lighting condition, and a relit image with a consistent mapping 
are typically synthesized using a renderer. \textcolor{black}{However, the distributional gap between 
synthetic and real data can limit the model's performance on real-world images. 
In particular, models trained exclusively on synthetic data may produce results 
with overly smooth textures and an artificial appearance, compromising photorealism.}

\begin{figure}[h]
    \centering
    \vspace{-2mm}
    \includegraphics[width=0.99\linewidth]{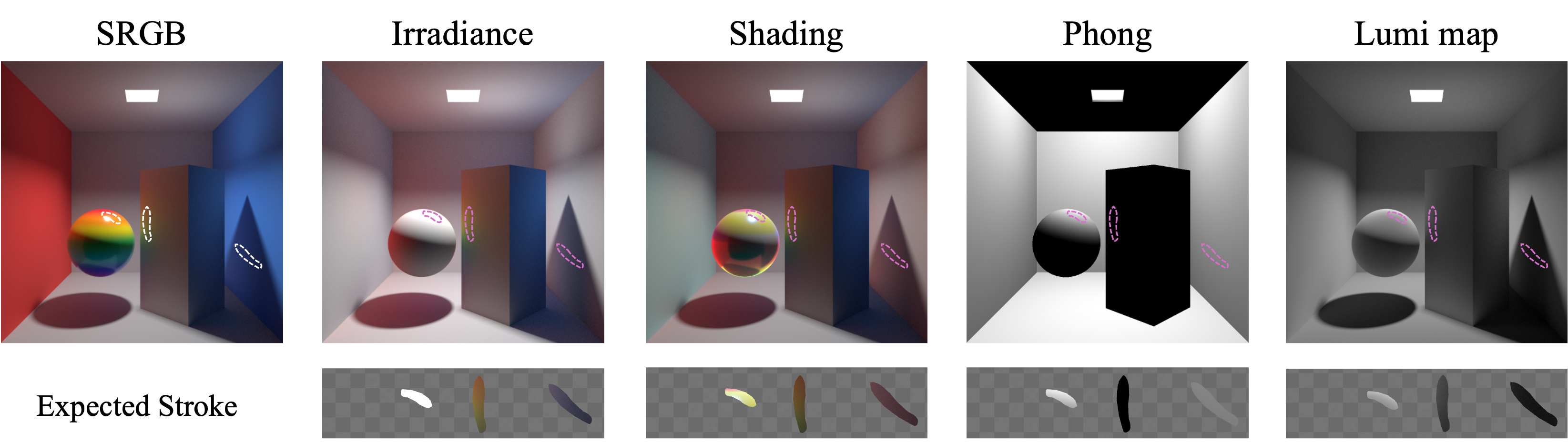}
    \caption{\textbf{Comparison of illumination representations.}
    In a Cornell box lit from above, irradiance 
    includes material-dependent interreflections, while shading ignores indirect illumination 
    and introduces additional errors through ideal Lambertian decomposition. Phong captures 
    only the light---normal relationship, omitting occlusion and distance attenuation. In contrast, 
    the Lumi Map directly represents the spatial distribution of source illumination incident 
    on scene surfaces.}
    \label{fig:representation}
\end{figure}

To address the first challenge, we define a material-decoupled illumination representation, 
termed the \textit{Lumi Map}, which encodes not only the spatial distribution of illumination 
but also its intensity and chromaticity. Existing methods typically adopt shading 
maps or Phong-based representations as lighting conditions. The former entangle 
illumination with the material properties of object surfaces, whereas the latter 
fail to capture occlusion relationships, as illustrated in Fig.~\ref{fig:representation}.
We learn the mapping between the lumi map and the corresponding relighting result. 
At inference time, modifying the values of the lumi map enables controllable relighting 
while preserving material consistency. Specifically, we carefully design a data-synthesis 
pipeline to ensure a consistent correspondence between each lumi map and its relighting 
result. By strictly controlling exposure parameters, normalizing material properties, and 
varying camera viewpoints and light-source attributes, we increase data diversity while 
preserving this mapping. Finally, we directly export the lumi map in linear color space 
and use it as the illumination condition.

For the second challenge, we train a lumi map estimator to infer lumi maps 
from real relighting image pairs, thereby constructing training triplets of the 
form \textcolor{black}{$(I_{src},L_{tar},I_{tar})$}, where \textcolor{black}{$I_{src}$} is the source image, \textcolor{black}{$L_{tar}$} is the target
illumination condition, and $I_t$ is the target image. Recent work, such as Vision
Banana~\citep{visionbanana}, has shown that generative models can learn visual representations through
fine-tuning with a relatively small amount of data. We realize this process through a two-stage training strategy. In the first stage, 
using synthetic data, we jointly train a general-purpose image editing model on two 
tasks: Lumi-driven image relighting and lumi map estimation from images. In the 
second stage, we incorporate real relighting pairs and estimate the illumination 
representation of the target image online during training, denoted as the target 
lighting \textcolor{black}{$L_{tar}$}. We then use \textcolor{black}{$L_{tar}$} as the conditioning signal, together with 
the source image \textcolor{black}{$I_{src}$}, to predict the target image \textcolor{black}{$I_{tar}$}. This design aligns the forward process
used in training with that used at inference time, while also improving the 
model's ability to generate realistic relit images.

We evaluate the accuracy of our method in performing relighting from sparse 
illumination representations on paired data with ground-truth targets. Both 
quantitative and qualitative results demonstrate the effectiveness of our 
approach. In addition, we evaluate its scribble-driven relighting performance 
on manually annotated real-world images. The experimental results show that 
our method achieves state-of-the-art relighting accuracy.

Our contributions can be summarized as follows:
(a) We introduce a material-decoupled illumination representation that facilitates learning a more consistent mapping between illumination conditions and relighting results, thereby enabling more accurate control.
(b) Building upon this high-quality illumination representation, we propose a unified training framework for relighting and illumination estimation, which extends training from synthetic data to real-world data.
(c) Finally, we present a state-of-the-art scribble-driven image relighting method that produces high-quality relighting results on real-world images.

\section{Related Work}
\label{sec:related}

\paragraph{Diffusion-based relighting.}

Recent methods leverage the image priors of diffusion models to improve relighting
realism~\citep{neuralgaffer,luminet} and combine them with diverse conditioning signals to provide different
degrees of control~\citep{dilightnet,lightit,lightlab,freelit}. IC-Light~\citep{iclight} scales diffusion-based illumination
editing to more than ten million training examples under a consistent-light-transport
constraint; its released models relight a supplied foreground using either text or a
background image as the condition. TokenLight~\citep{tokenlight} encodes lighting
attributes---including source position, chromaticity, intensity, ambient illumination,
and diffuse level---as tokens, enabling parameter-based control of the resulting
illumination. Lume-Palette~\citep{lumepalette} focuses on spatial consistency in
multi-view indoor relighting: it derives dense spatial lighting conditions from coarse
reconstructed geometry and uses them to cast canonical illumination priors.
General-purpose multimodal generators, such as Qwen-Image-Edit~\citep{qwenedit} and
Nano Banana 2~\citep{nanobanana2}, also support relighting as an image-editing task
conditioned on input images and textual instructions. Although these methods explore
how additional conditions can improve relighting control, practical controllability
remains limited by the difficulty of providing conditioning signals that are both
easy to author and sufficiently precise.

\paragraph{Scribble-guided relighting.}

Digital art shows that drawing can specify illumination directly in image space~\citep{illuminationbrush}.
SmartShadow~\citep{smartshadow} is an early data-driven system that converts sparse
2D marks into artistic shadows.
Although limited to line drawings rather than photorealistic relighting, it establishes
such marks as an intuitive illumination interface. LightPainter~\citep{lightpainter}
enables interactive portrait relighting from freehand shading strokes. It represents
surface illumination with Phong shading and samples training strokes from quantized
Phong maps. Because Phong omits visibility and distance attenuation, these strokes
cannot describe complex spatial light transport; the method's portrait-specific
design further limits its use on unconstrained photographs.
ScribbleLight~\citep{scribblelight} extends scribble guidance to indoor scenes using
albedo-conditioned diffusion and a ControlNet~\citep{controlnet} with predicted normals and
binary brighten--darken annotations. Its strokes serve as binary labels for bright and
shadowed regions, encoding neither continuous relative magnitude nor chromaticity.
Unlike these methods, our material-decoupled representation retains spatial support,
relative magnitude, and chromaticity per sparse sample, enabling fine-grained
brightness and color control.

\paragraph{Illumination representations for image decomposition.}

Prior work exposes illumination at different levels of abstraction. The 
Phong model~\citep{phong1975} expresses appearance as a combination of ambient, 
diffuse, and specular terms, but its coefficients conflate illumination 
with material response. Intrinsic-image decomposition~\citep{barrow1978,intrinsic} 
factors an observed image into reflectance and shading. Although shading 
provides an editable image-space layer, it remains coupled with scene geometry 
and depends on decomposition accuracy. Irradiance offers a more physically 
grounded representation, defined as the cosine-weighted integral of incident 
radiance over the hemisphere at a surface. Under distant illumination, diffuse 
irradiance is dominated by low-frequency components~\citep{ramamoorthi2001}. 
Although independent of the local receiving material, irradiance may include 
indirect illumination and therefore depend on materials elsewhere in the scene.
Recent diffusion models expose such decomposed quantities as editing controls. 
RGB$\leftrightarrow$X~\citep{rgbx} estimates albedo, normals, roughness, metallicity, 
and diffuse irradiance from an RGB image and synthesizes images from all or a subset 
of these channels. Diffusion Renderer~\citep{diffusionrenderer} estimates editable 
G-buffers, including albedo, normals, depth, roughness, and metallicity, from video 
and conditions a learned renderer on these buffers and an environment map. These 
factorized interfaces support relighting and material editing but require dense 
intrinsic estimates and explicit lighting inputs, rather than sparse image-space observations.

\section{Illumination Representation with Material Decoupling}
\label{sec:rep}

\begin{wraptable}{r}{0.5\linewidth}
  \vspace{-3mm}
  \begin{minipage}{\linewidth}
  \caption{\textbf{Notation for light transport.}}
  \label{tab:notation}
  \centering
  \footnotesize
  \setlength{\tabcolsep}{4pt}
  \renewcommand{\arraystretch}{1.08}
  \begin{tabular}{@{}>{\raggedright\arraybackslash}m{0.30\linewidth}
                      >{\raggedright\arraybackslash}m{\dimexpr0.70\linewidth-2\tabcolsep\relax}@{}}
  \toprule
  \textbf{Symbol} & \textbf{Meaning} \\
  \midrule
  $\mathcal{G},\mathcal{M},\Lambda,\mathcal{C}$ & Geometry, materials, \textcolor{black}{light source parameters}, and camera. \\
  $x,n_x$ & Surface point and its geometric normal. \\
  $\omega_i,\omega_o,\mathcal{H}_x$ & Incident/outgoing directions and the upper hemisphere at $x$. \\
  $L_i,L_o,L_e$ & Incident, outgoing, and emitted radiance. \\
  $L_{i,\mathrm{src}},L_{i,\mathrm{ind}}$ & Direct-source and indirectly reflected incident radiance. \\
  $f_r^{\mathcal{M}}$ & BRDF determined by the original materials. \\
  $p,x_p$ & Pixel and its first surface intersection. \\
  $I_s,I_t,L_s,L_t$ & Source/target images and their lumi-map conditions. \\
  \bottomrule
  \end{tabular}
  \end{minipage}
  \end{wraptable}
A sparse lighting stroke is useful only if its retained value has a predictable effect
on the completed image. Table~\ref{tab:notation} summarizes the symbols used below. For fixed geometry
$\mathcal{G}$, material field $\mathcal{M}$, \textcolor{black}{light source parameters}
$\Lambda$, and camera $\mathcal{C}$, we seek a dense signal that can later be
sparsified while retaining image-space support, relative magnitude, and chromaticity.
The signal should exclude the original material response but retain the geometric
factors that place direct illumination and cast shadows. We derive it under the
standard steady-state, surface-only light-transport model: radiance is conserved along
unobstructed rays, no participating medium is present, and scattering vertices are
described by opaque BRDFs. These assumptions match the reflected component used by our
canonical direct-diffuse pass. All radiance quantities below may be evaluated
spectrally or per RGB channel. We follow the standard convention that both incident
and outgoing directions point away from their local surface.

\paragraph{Full light transport.}
At a \textcolor{black}{non-emissive} surface point $x$, energy balance gives the classical rendering
equation~\citep{kajiya1986,pbrt4}:
\begin{equation}
\label{eq:render}
\begin{aligned}
L_o^{\mathcal{G},\mathcal{M},\Lambda}
(x,\omega_o)
={}&\int_{\mathcal{H}_x}
\underbrace{f_r^{\mathcal{M}}(x,\omega_i,\omega_o)}_{\text{material coupling}}
L_{i,\mathrm{src}}^{\mathcal{G},\Lambda}
(x,\omega_i)
(n_x\!\cdot\!\omega_i)_+
\,\mathrm{d}\omega_i \\
&+\underbrace{\int_{\mathcal{H}_x}
f_r^{\mathcal{M}}(x,\omega_i,\omega_o)
L_{i,\mathrm{ind}}^{\mathcal{G},\mathcal{M},\Lambda}(x,\omega_i)
(n_x\!\cdot\!\omega_i)_+
\,\mathrm{d}\omega_i}_{\text{indirect reflection}} .
\end{aligned}
\end{equation}
Here $(a)_+=\max\{a,0\}$. Equation~\eqref{eq:render} is complete within the stated
surface model: $L_i$ includes source emission as well as radiance that has undergone
any number of earlier reflections. To expose the remaining material dependence, we
partition the incident field into radiance arriving directly from a source and
radiance arriving after at least one surface reflection:$L_i^{\mathcal{G},\mathcal{M},\Lambda}=L_{i,\mathrm{src}}^{\mathcal{G},\Lambda}
+L_{i,\mathrm{ind}}^{\mathcal{G},\mathcal{M},\Lambda}.$

\paragraph{Canonicalizing the receiver material.}
We replace the BRDF at the receiver by a white, unit-albedo Lambertian constant,
$f_r^{\mathcal{M}}\!\rightarrow f_r^{\mathrm{can}}
=\rho^{\mathrm{can}}/\pi=1/\pi$ with $\rho^{\mathrm{can}}=1$.
Substitution into Eq.~\eqref{eq:render} gives
\begin{equation}
\label{eq:canonical-render}
\begin{aligned}
L_{o,\mathrm{can}}^{\mathcal{G},\mathcal{M},\Lambda}
(x,\omega_o)
={}&\frac{1}{\pi}\int_{\mathcal{H}_x}
\Bigl[L_{i,\mathrm{src}}^{\mathcal{G},\Lambda}
(x,\omega_i)
+L_{i,\mathrm{ind}}^{\mathcal{G},\mathcal{M},\Lambda}
(x,\omega_i)\Bigr]
(n_x\!\cdot\!\omega_i)_+
\,\mathrm{d}\omega_i .
\end{aligned}
\end{equation}
This canonicalization removes only the receiver's original BRDF. The indirect field
$L_{i,\mathrm{ind}}$ still depends on the materials of earlier scattering surfaces, so
canonicalizing the receiver alone does not yet produce a material-independent signal.

\paragraph{Removing material-dependent interreflection.}
The lumi map deliberately retains only the direct-source component of
Eq.~\eqref{eq:canonical-render}; it is not asserted to approximate the complete image.
For pixel $p$, let $x_p=x_{\mathcal{G},\mathcal{C}}(p)$ be the first
surface intersection under camera $\mathcal{C}$. We define
\begin{equation}
\label{eq:diffdir}
\begin{aligned}
L_t(p)
&\equiv L_{o,\mathrm{can}}^{\mathrm{dir}}
(x_p,\omega_o) \\
&=\frac{1}{\pi}\int_{\mathcal{H}_{x_p}}
L_{i,\mathrm{src}}^{\mathcal{G},\Lambda_t}
(x_p,\omega_i)
(n_{x_p}\!\cdot\!\omega_i)_+
\,\mathrm{d}\omega_i .
\end{aligned}
\end{equation}
Together, the source field, its directional support, and the receiver cosine encode
source magnitude and chromaticity as well as geometry-dependent visibility,
orientation, and distance coupling. Consequently, Eq.~\eqref{eq:diffdir} depends on
$\mathcal{G}$, $\Lambda_t$, and $\mathcal{C}$, but contains no factor from the original
material field $\mathcal{M}$. It preserves the spatial support of direct illumination
and cast shadows while excluding both the receiver response and material-colored
interreflection.

In practice, we replace every non-emitting surface material by the same white,
unit-albedo matte and extract the direct-diffuse component. The resulting RGB
condition $L_t$ is pixel-aligned with the target image $I_t$: visibility and attenuation
define the support of light and shadow, while per-channel values encode relative
magnitude and chromaticity. These values are relative control signals rather than
calibrated radiometric measurements. Equation~\eqref{eq:diffdir} shows that this 
representation provides a more direct interface for illumination control and simplifies 
the construction of conditioning inputs at inference time.

\section{Method}
\label{sec:method}

In this section, we describe how to train a hand-drawn relighting model using the 
lumi map defined in Section \ref{sec:rep}. Due to space constraints, we briefly describe Lumi Map synthesis and sparsification in 
Supplementary~\ref{sec:method:control} and our two-stage training strategy in Section~\ref{sec:method:model}.

\subsection{From Dense Rendering Supervision to Sparse Strokes}
\label{sec:method:data}

\begin{figure}[thb]
  \centering
  \includegraphics[width=0.9\linewidth]{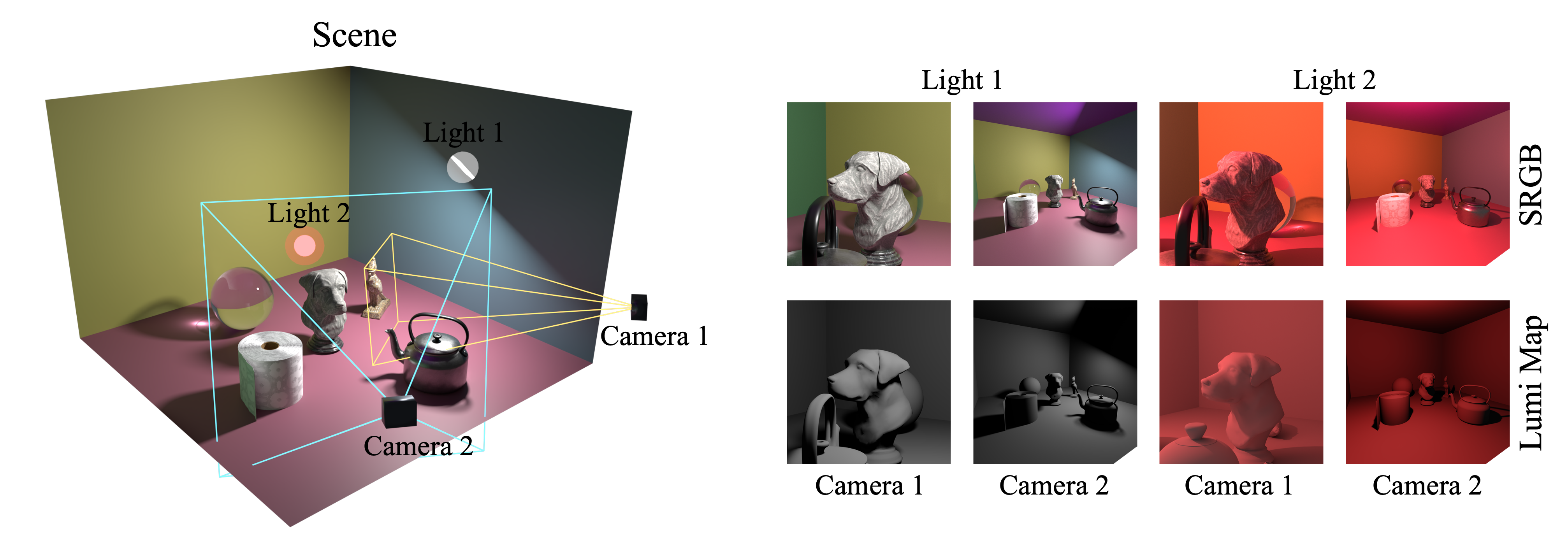}
  \caption{\textbf{Construction of paired rendered supervision.} A scene is sampled
  under multiple cameras and \textcolor{black}{varying light source parameters} (\emph{left}). Each configuration
  yields a rendered sRGB image (\emph{top}) and a pixel-aligned,
  canonical-material lumi map (\emph{bottom}); both share the
  same geometry, viewpoint, and source.}
  \label{fig:method}
  \end{figure}

\paragraph{Paired variation in source position and chromaticity.}
The rendering pipeline is organized as scene, lighting condition, and camera
(Figure~\ref{fig:method}). Its scenes include
isolated subjects, interacting objects, and interiors so that illumination produces
self-shadowing, cast shadows, and foreground--background interactions. For each
fixed scene and view, we vary the source position and chromaticity. Monochromatic
basis renders are combined in linear-radiance space to cover colored illumination
without changing geometry or material.

\paragraph{Pixel-aligned supervision.}
Camera and source directions cover the scene, and configurations with a visible
emitter are filtered out. For each configuration, one pass renders the image with its
original materials and a second pass renders Eq.~\eqref{eq:diffdir} with the canonical
material. The shared scene, camera, and source establish pixelwise
correspondence between the target image and its lumi map. The synthetic training and
validation sets are split by scene, excluding validation geometry from synthetic
training. During data synthesis, we keep the exposure settings fixed and dynamically
vary the light-source intensity to generate diverse relighting results.

\vspace{-2mm}
\begin{figure}[th]
    \centering
    \includegraphics[width=0.99\linewidth]{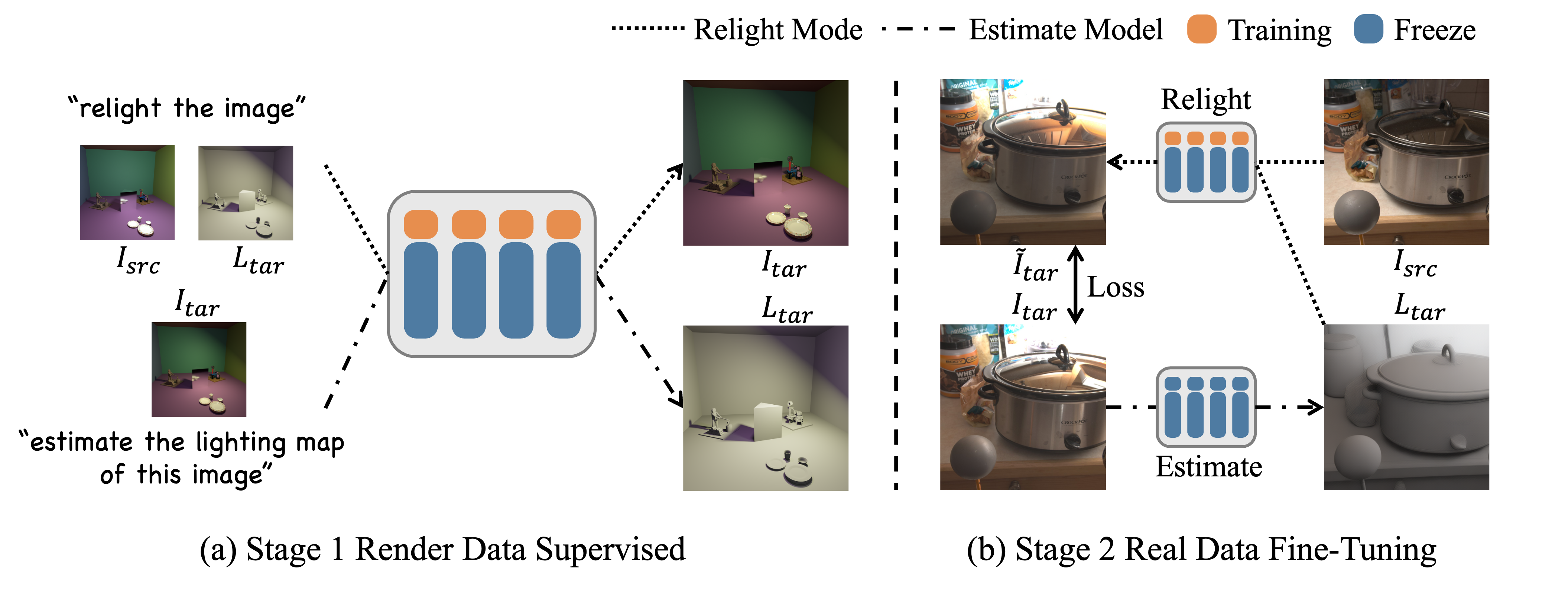}
    \caption{\textbf{Two-stage training strategy.}
    \emph{Stage 1} optimizes a shared network for relighting from a rendered lumi map
    and for estimating that map from an image. \emph{Stage 2} adds paired real-image
    reconstruction using a target-derived pseudo-control with stopped gradients, while
    rendered supervision remains active. \textcolor{black}{The figure illustrates the case of a dense 
    lumi map, whereas training uses lumi maps with randomly sampled sparsity levels.}}
    \label{fig:training}
    \end{figure}

\subsection{Two-Stage Training with Rendered Supervision and Real Reconstruction}
\label{sec:method:model}

\paragraph{Stage 1: learning dense illumination and sparse completion.}
We denote the source and target images by $I_{src}$ and $I_{tar}$, and their corresponding
lumi-map illumination conditions by $L_{src}$ and $L_{tar}$. We train the DiT~\citep{dit} model in
latent space, where each image $I$is encoded into a latent $z=\mathcal{E}(I)$.
On the dense and sparsified rendered pairs of Section~\ref{sec:method:data}, we train with the standard flow-matching
objective~\citep{flowmatching,rectifiedflow}. Let
the clean latent--condition pairs for mode $m\in\{G,E\}$ be
$(x_0^G,c_G)=(z_{tar},(z_{src},z_{L_{tar}}))$ for relighting and
$(x_0^E,c_E)=(z_{L_{src}},z_{src})$ for estimation. The mode $m$ is controlled by varying the 
editing prompt, as shown in Fig.~\ref{fig:training}.With timestep
$t\sim\mathcal{U}(0,1)$ and
$\epsilon\sim\mathcal{N}(0,I)$, let
$x_t^m=(1-t)x_0^m+t\epsilon$ and
$v^m=\mathrm{d}z_t^m/\mathrm{d}t=\epsilon-x_0^m$. Stage~1 optimizes
\begin{equation}
\label{eq:loss}
\mathcal{L}_{\mathrm{flow}}(\theta)
=\mathbb{E}_{m,\,c,\,\epsilon,\,t}
\left\|v_{\theta}(x_t^m,t,c_m)-v^m\right\|^2,
\end{equation}
so the same parameters learn relighting $(I_{src},L_{tar}\!\to\!I_{tar})$ and lumi estimation
$(I_{tar}\!\to\!L_{tar})$ under their respective mode labels.

\paragraph{Stage 2: estimator-guided reconstruction on real photos.}
Training only on rendered data can bias the model toward synthetic appearances and
degrade photorealistic generation. We therefore incorporate real relighting
pairs~\citep{multiillum,rene}, which lack ground-truth target illumination conditions
$L_{tar}$. For each pair
$(I_{src},I_{tar})$, the target image supplies a pseudo-control that
conditions relighting of the source. Each real-data optimization step first estimates
this condition using the estimation mode:
\begin{equation}
\label{eq:estimate-real}
\widetilde{z}_{L_{tar}}=\operatorname{sg}\!\left[E_\theta(z_{I_{tar}})\right],
\end{equation}
where $E_\theta$ denotes the model's forward pass in estimation mode and $\operatorname{sg}$ denotes the stop-gradient operator. 
The reconstruction loss is then defined as:
\begin{equation}
\label{eq:recon}
\mathcal{L}_{\mathrm{rec}}(\theta)
  = \mathbb{E}_{z_{src},\,\widetilde{z}_{L_{tar}},\,\epsilon,\,t}
  \left\|v_{\theta,G}(x_t^G,t,z_{src},\widetilde{z}_{L_{tar}})-v^G\right\|^2.
\end{equation}
Using only the reconstruction loss conditioned on estimated illumination may expose 
the model to erroneous inputs when the estimated condition is inaccurate, potentially 
degrading its performance. Therefore, in Stage 2, we additionally use synthetic data 
to jointly train the model in both relighting and estimation modes. The overall 
training objective is defined as:
\begin{equation}
  \label{eq:total_loss}
  \mathcal{L}(\theta)=\mathcal{L}_{\mathrm{flow}}(\theta)+\mathcal{L}_{\mathrm{rec}}(\theta)
  \end{equation}

\section{Experiments}
\label{sec:exp}

\paragraph{Implementation details.}
We use Qwen-Edit-2511\cite{qwenedit} as the backbone and fine-tune it with LoRA~\citep{lora} using a rank of 64.
For render data, we construct 4.7K scenes in Blender and render 115K images under 
diverse lighting conditions. For real-world relighting data, we collect 36K images 
from MultiIllum\cite{multiillum} and ReNe\cite{rene}. In Stage 1, the model is trained for 15K steps at a 
resolution of $512 \times 512$ with a learning rate of $1\times10^{-4}$. 
In Stage 2, it is trained for an additional 5K steps at a resolution of $1024 \times 1024$ 
with a learning rate of $5\times10^{-5}$. We use the AdamW~\citep{adamw} optimizer. During inference,
the classifier-free guidance~\citep{cfg} scale is set to 4, and sampling steps is set to 40.

\paragraph{Benchmarks and metrics.}
We construct three benchmarks---ReNe~\citep{rene}, MultiIllum~\citep{multiillum}, and SynthBench---by selecting held-out 
image pairs with pronounced illumination changes, with 600 pairs in each benchmark. 
The three benchmarks exhibit progressively increasing lighting complexity. 
ReNe contains fixed scenes with a single subject relit by varying the position 
of a white light source. MultiIllum includes more diverse content but still uses 
a single white light source. SynthBench offers the greatest diversity, covering 
various relighting scenarios, including portraits and complex scenes. We 
additionally collect 64 photographic images from the Internet as an in-the-wild 
test set to evaluate generalization. Following prior work, we measure similarity 
to the ground truth using RMSE, PSNR, SSIM~\citep{ssim}, and LPIPS~\citep{lpips}.

\paragraph{Baselines.}
We select ScribbleLight~\citep{scribblelight}, a scribble-driven relighting method, as a baseline, and 
additionally include IC-Light~\citep{iclight} and multimodal image editing models, including Banana~\citep{nanobanana2}, 
Qwen-Edit~\citep{qwenedit}. To ensure a fair comparison, each baseline is evaluated using 
the conditioning format best suited to its design. Due to space constraints, we limit comparisons in the 
main paper to the most relevant methods or those with the best overall performance. Additional comparisons are provided in the Supplementary \ref{app:dense-baselines}.

\subsection{Comparison with Existing Methods}
\label{sec:exp:comparison}

\begin{figure}[!ht]
        \centering
        \includegraphics[width=0.99\linewidth]{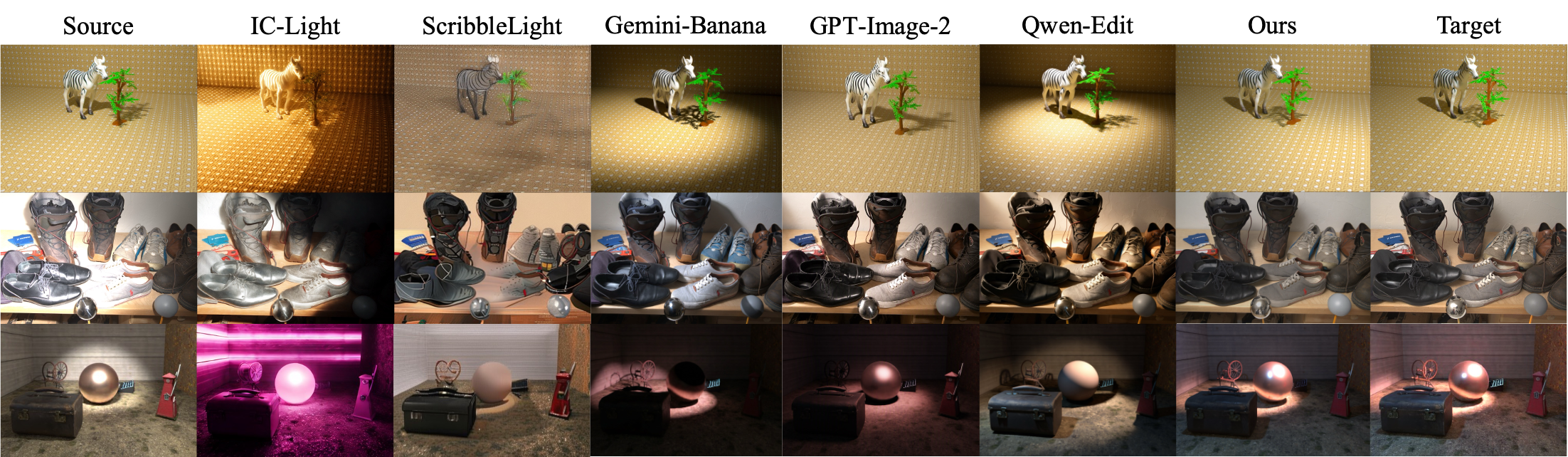}
        \caption{Qualitative comparison of relighting accuracy with existing methods across 3 benchmarks.}
        \label{fig:qualitative}
        \end{figure}
        
        \begin{figure}[!htb]
                \centering
                \includegraphics[width=0.99\linewidth]{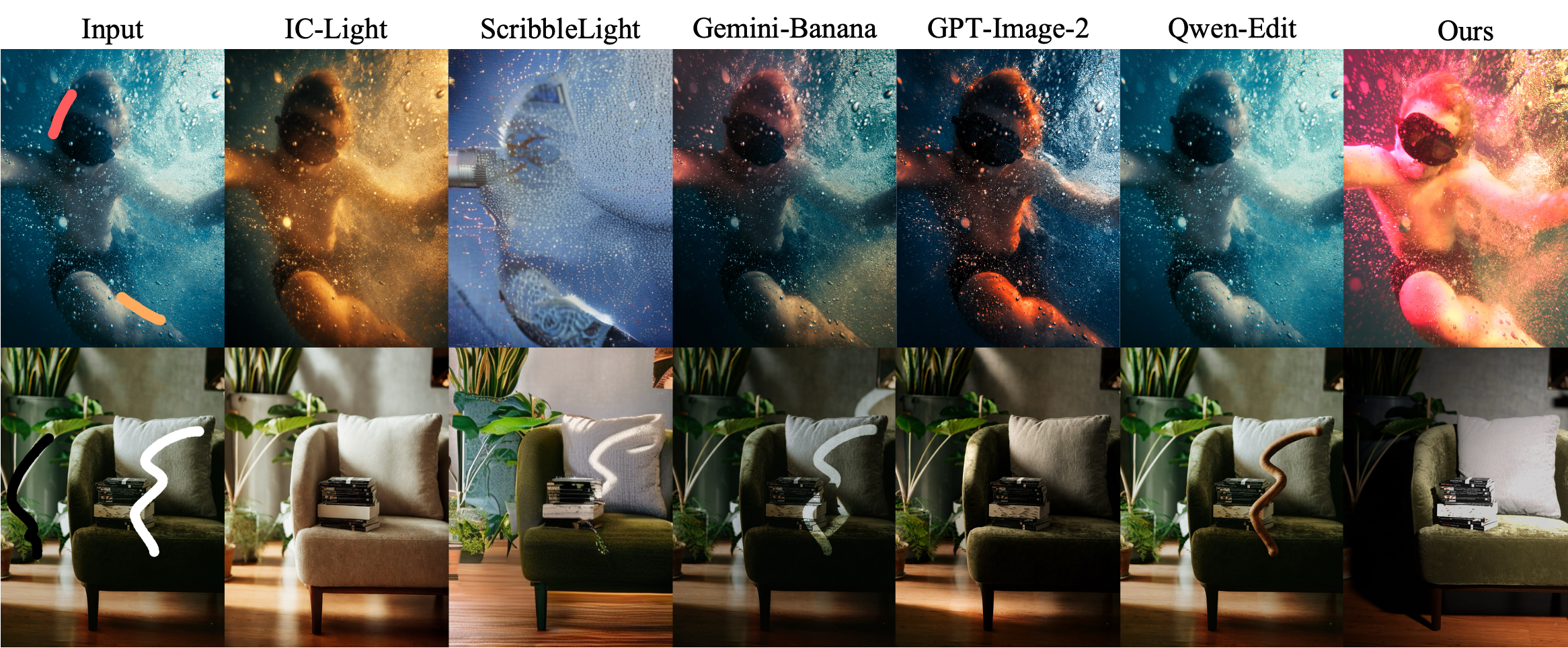}
                \caption{Visual analysis of relighting results on in-the-wild images. }
                \label{fig:qualitative-itw}
                \end{figure}

\paragraph{Qualitative analysis.} Figure~\ref{fig:qualitative} presents a visual comparison between the 
relighting results of different methods and the corresponding ground truth. The first row shows results 
on the ReNe benchmark, where our method more accurately reproduces the locations of shadows and highlights, 
as well as the target illumination intensity. The second row presents results on MultiIllum, demonstrating 
that our method better preserves object materials while accurately matching the illumination chromaticity.
The third row evaluates relighting in complex synthetic scenes, where competing methods fail to 
reproduce the target illumination due to insufficient control, whereas dior achieves accurate relighting.

We present qualitative results on the in-the-wild dataset in Fig.~\ref{fig:qualitative-itw}. The first row 
shows that our method remains responsive to scribble guidance and produces realistic relighting results 
even in challenging cases, such as underwater scenes. In contrast, existing methods either fail to generate 
illumination with the desired direction and color or produce low-quality images. The second row further 
highlights the advantage of our method in terms of physical consistency, for example, in reproducing shadows 
cast by strong illumination and distinct specular highlights on surfaces of different materials.

\FloatBarrier
\begin{table}[!ht]
\centering
\caption{Quantitative comparison and ablation across three benchmarks.}
\footnotesize
\setlength{\tabcolsep}{3pt}
\resizebox{\linewidth}{!}{%
\begin{tabular}{l cccc cccc cccc}
\toprule
& \multicolumn{4}{c}{\textbf{ReNe}}
& \multicolumn{4}{c}{\textbf{MultiIllum}}
& \multicolumn{4}{c}{\textbf{SynthBench}} \\
\cmidrule(lr){2-5}\cmidrule(lr){6-9}\cmidrule(lr){10-13}
Method & RMSE$\downarrow$ & PSNR$\uparrow$ & SSIM$\uparrow$ & LPIPS$\downarrow$
       & RMSE$\downarrow$ & PSNR$\uparrow$ & SSIM$\uparrow$ & LPIPS$\downarrow$
       & RMSE$\downarrow$ & PSNR$\uparrow$ & SSIM$\uparrow$ & LPIPS$\downarrow$ \\
\midrule
IC-Light
& 86.52 & ~9.47 & 0.3450 & 0.4783 & 96.83 & ~8.70 & 0.3753 & 0.5742 & 85.02 & ~9.74 & 0.3425 & 0.5850 \\
ScribbleLight
& \underline{47.46} & \underline{14.64} & 0.2819 & 0.5632
& \underline{56.29} & \underline{13.39} & 0.5355 & 0.5834
& 76.10 & 10.90 & 0.3945 & 0.5356 \\
Gemini-Banana
& 59.37 & 13.17 & \underline{0.4576} & \underline{0.3319}
& 61.98 & 13.08 & 0.5798 & 0.3554
& \underline{57.42} & \underline{14.11} & 0.5581 & \underline{0.3192} \\
GPT-Image-2
& 47.89 & 14.64 & 0.2320 & 0.4256
& 66.53 & 12.38 & 0.5663 & \underline{0.3496}
& 62.37 & 12.94 & 0.4902 & 0.3420 \\
Qwen-Image-Edit
& 50.16 & 14.45 & 0.4350 & 0.3483 & 64.64 & 12.50 & \underline{0.5968} & 0.3677 & 57.48 & 13.86 & \underline{0.5771} & 0.3256 \\
\midrule
\rowcolor{black!8}
w Irradiance
& \underline{24.16} & \underline{20.93} & \underline{0.7333} & \underline{0.1942}
& \underline{36.41} & \underline{17.54} & \underline{0.7602} & \underline{0.2410}
& 32.62 & 19.10 & 0.7974 & 0.1920 \\
\rowcolor{black!8}
w Shading
& 37.95 & 16.74 & 0.7089 & 0.2537
& 38.76 & 16.78 & 0.7483 & 0.2554
& 35.89 & 18.29 & 0.7563 & 0.2134 \\
\rowcolor{black!8}
w Phong
& 39.66 & 16.64 & 0.6858 & 0.2402
& 51.51 & 14.59 & 0.7026 & 0.2964
& 49.94 & 14.67 & 0.7778 & 0.2441 \\
\rowcolor{black!8}
Dior w/o Stage~2
& 25.09 & 20.34 & 0.7321 & 0.2195
& 45.86 & 15.65 & 0.6947 & 0.2871
& \underline{29.82} & \underline{19.69} & \underline{0.8293} & \underline{0.1656} \\
\midrule
\textbf{\method{} (ours)}
& \textbf{20.03} & \textbf{22.38} & \textbf{0.7542} & \textbf{0.1740} & \textbf{33.49} & \textbf{18.21} & \textbf{0.7659} & \textbf{0.2275} & \textbf{24.89} & \textbf{21.12} & \textbf{0.8637} & \textbf{0.1166} \\
\bottomrule
\end{tabular}}
\label{tab:main}
\label{tab:repr-sparse}
\end{table}

\begin{wraptable}{r}{0.35\linewidth}
        \vspace{-3mm}
        \begin{minipage}{\linewidth}
        \centering
        \caption{Elo ratings on itw images.}
        \label{tab:itw-elo}
        \scriptsize
        \setlength{\tabcolsep}{2pt}
        \resizebox{\linewidth}{!}{%
        \begin{tabular}{@{}lrr@{}}
        \toprule
        Method / variant & Elo$\uparrow$ & W/L/T (\%) \\
        \midrule
        IC-Light & 877.99 & 41.1/58.9/0.0 \\
        ScribbleLight & 409.56 & 4.2/95.8/0.0 \\
        Gemini-Banana & 1169.55 & 63.4/36.6/0.0 \\
        GPT-Image-2 & 1113.67 & 53.6/46.1/0.3 \\
        Qwen-Image-Edit & 661.25 & 16.3/83.7/0.0 \\
        \midrule
        w Irradiance & 1320.85 & 82.2/17.8/0.0 \\
        w Shading & 1109.59 & 68.3/31.7/0.0 \\
        w Phong & 953.00 & 54.9/45.1/0.0 \\
        w/o stage2 & 1181.85 & 70.6/29.4/0.0 \\
        \midrule
        \textbf{\method{} (ours)} & \textbf{1410.65} & \textbf{87.2}/12.8/0.0 \\
        \bottomrule
        \end{tabular}}
        \end{minipage}
        \vspace{-2mm}
        \end{wraptable}
\paragraph{Quantitative analysis.} Table~\ref{tab:main} compares the relighting 
accuracy of our method with existing approaches. Existing methods provide 
limited illumination control, resulting in substantial discrepancies from the 
target relit images. In contrast, our method enables effective illumination 
control and achieves the best performance across all evaluation settings.

To quantitatively evaluate relighting on in-the-wild images without ground 
truth, we conduct randomized pairwise comparisons of outputs from different 
methods, using GPT-5.2 as a vision–language model (VLM) judge. The VLM selects 
the preferred result based on three criteria: adherence to the input 
conditions, light-transport consistency, and photorealism. We compute Elo 
ratings from approximately $2,000$ comparisons. Our method achieves the best 
performance, with a win rate of $87.2\%$ and an Elo rating of $1410.65$. We 
emphasize that this evaluation is not a rigorous assessment of physical 
simulation accuracy but serves as a proxy for user experience in practical 
applications. Further details are provided in Supplementary~\ref{app:itw-elo}.
\FloatBarrier

\subsection{Ablation Study}
\label{sec:exp:ablation}
\label{sec:exp:repr}

Table~\ref{tab:repr-sparse} reports their impact on relighting accuracy. We repeat training with 
different illumination representations and evaluate relighting performance using the corresponding 
sparse inputs at inference time. Replacing the 
Lumi Map with alternative illumination representations consistently degrades performance. 
The gap from the full model is relatively small on ReNe and MultiIllum, where the 
light-source attributes are comparatively simple, but becomes substantially larger on 
SynthBench, which contains more complex illumination. This result highlights the advantage 
of the material-decoupled design of the Lumi Map at inference time. Omitting the Stage 2 training 
strategy while keeping the total training budget unchanged also leads to a clear performance drop 
on real-image benchmarks, demonstrating the effectiveness of reconstruction training with real 
relighting pairs.

\begin{figure}[h]
\centering
\includegraphics[width=\linewidth]{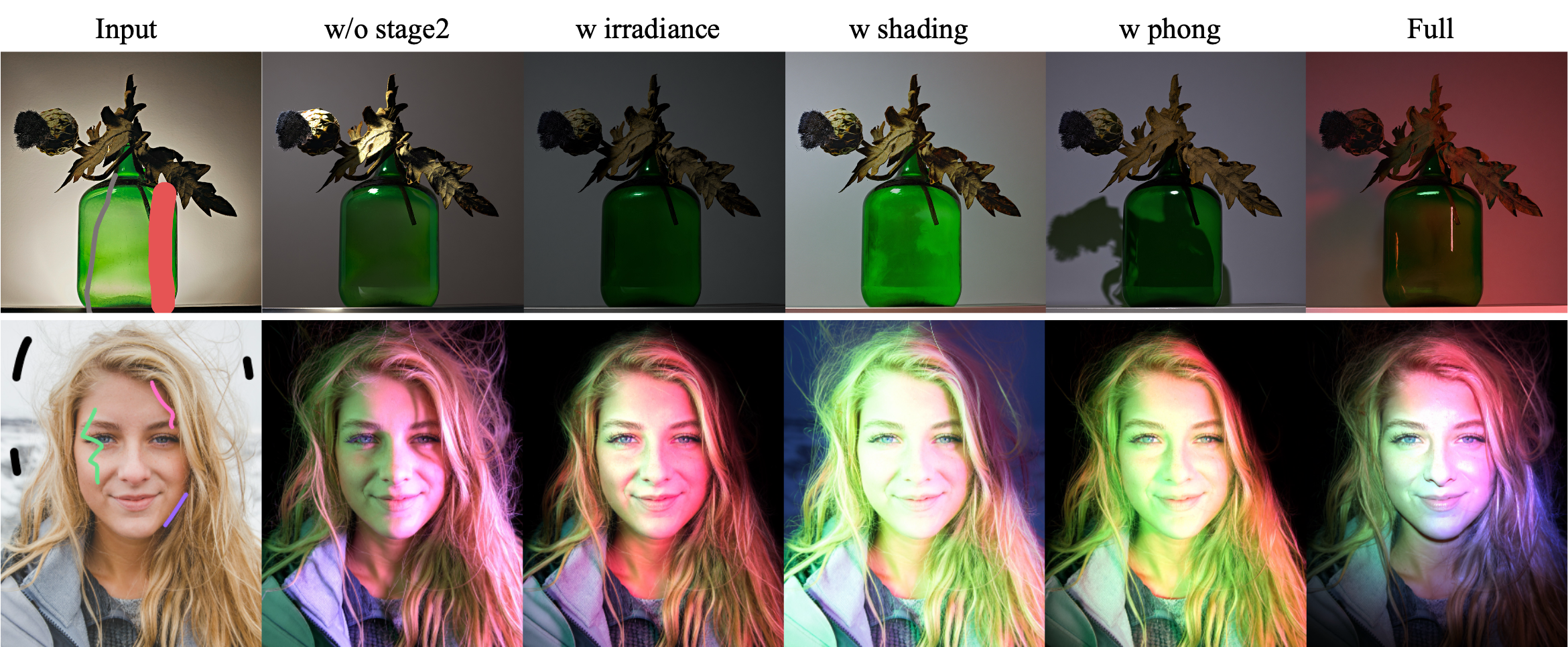}
\caption{\textbf{Qualitative ablation of illumination representations and Stage~2.}
Each row uses the same source photograph and sparse lumi drawing for the three
alternative representations, Dior w/o Stage~2, and the final Dior model. The cases
compare chromatic and spatial responses, respectively. No relit reference is
available, so the figure is not used for quantitative ranking.}
\label{fig:repr-response}
\end{figure}

Figure~\ref{fig:repr-response} presents the corresponding qualitative ablations. When trained with illumination 
representations other than the Lumi Map, the model follows the lighting scribbles less reliably, 
as shown in the first row. Shading additionally causes noticeable color shifts, such as the 
oversaturated facial appearance in the second row. Without Stage 2 training, the relighting 
results appear less natural, with abrupt illumination transitions across the face and inconsistent 
blending between the foreground and the black background.

\vspace{-2mm}
\FloatBarrier
\vspace{-3mm}
\subsection{Additional analysis for application}
\label{sec:exp:application}

\begin{figure}[!ht]
\centering
\includegraphics[width=0.99\linewidth]{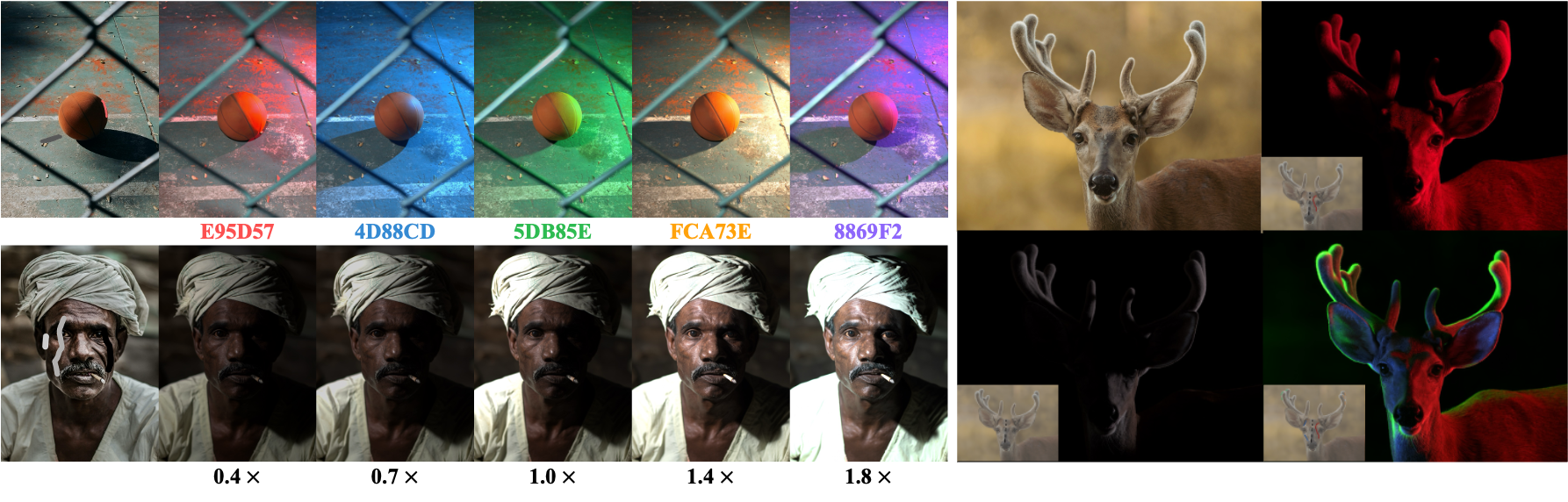}
\vspace{-2mm}
\caption{Visualization of illumination intensity and chromaticity control with our method.
}
\label{fig:fine-control}
\end{figure}

In this section, we demonstrate the applicability of our method to practical
relighting scenarios. As shown in Figure~\ref{fig:fine-control}, users can adjust the chromaticity and intensity of the
illumination to achieve customized lighting effects. They can also progressively 
add strokes to obtain more refined relighting results. Overall, our method 
enables users to create complex relighting effects through intuitive hand-drawn strokes.

\FloatBarrier

\vspace{-4pt}

\vspace{-4pt}
\section{Conclusion}
\label{sec:conclusion}
\vspace{-2pt}

In this work, we presented \textbf{Dior-Light}, a hand-drawn image relighting method
built upon the \textbf{Lumi Map}, a material-decoupled representation that encodes
the spatial distribution, relative intensity, and chromaticity of direct 
illumination. We constructed pixel-aligned synthetic training triplets and 
adopted a two-stage training strategy that combines rendered supervision with 
estimator-guided reconstruction on real relighting pairs. Experiments on three 
paired benchmarks show that the proposed representation and training strategy 
improve relighting accuracy over the evaluated baselines, while qualitative results 
indicate that the model can respond consistently to sparse hand-drawn controls 
on in-the-wild images. Overall, Dior-Light provides a practical approach to controlling 
illumination distribution, intensity, and chromaticity without explicit geometry 
or material estimation or per-image optimization.   

\clearpage
\section*{AI use statement}
We used AI tools for translation, language polishing, and grammar correction. The authors reviewed
and take responsibility for all content.

\bibliographystyle{iclr2027_conference}
\bibliography{refs}

\clearpage
\appendix
\section*{Appendix}

\noindent\textbf{Discussion of closely related work.} LightIt~\citep{lightit},
DiLightNet~\citep{dilightnet}, and Lume-Palette~\citep{lumepalette} share our use of
explicit illumination representations. LightIt uses direct shading that includes cast
shadows as a diffusion conditioning signal and trains a shading estimator to construct
lighting conditions for real-image training data. DiLightNet renders diffuse and specular
radiance hints with homogeneous proxy materials, accounting for shadows and indirect
illumination. Lume-Palette similarly renders scenes with a uniform, untextured white
material to obtain spatial lighting conditions for multi-view indoor relighting. These
methods provide closely related formulations, but their representations and interaction
mechanisms are not designed specifically for sparse, hand-drawn lighting control.

Our work instead investigates how the choice of illumination representation affects
relighting accuracy and controllability under sparse hand-drawn guidance at inference
time. We compare several existing representations, formulate the material-decoupled Lumi
Map based on the rendering equation, and validate its effectiveness through ablation
studies. The Lumi Map is designed as a user-editable lighting interface, with stroke
locations and values specifying the spatial distribution, intensity, and chromaticity of
illumination. We further develop a unified relighting and estimation training strategy
that enables the model to respond to these sparse conditions. Thus, although these works
share the use of illumination representations, they differ in their design objectives and
application settings. Our contribution lies in an illumination representation tailored to
sparse, user-specified control at inference time, together with a training strategy that
supports this interaction.

\clearpage

\section{Lumi Representation Details}
\label{app:reps}

\subsection{Surface-Area Expansion}
\label{app:transport}

We give surface-area forms of Eqs.~\eqref{eq:render}--\eqref{eq:diffdir},
following the construction in Section~\ref{sec:rep}: full transport, a canonical
receiver under the original incident field, and direct-source transport.
We assume steady-state transport, opaque surfaces, and no participating media.
The derivation uses finite area emitters; ideal point sources are given separately.

\paragraph{Geometry and transport kernel.}
Let $\mathcal{A}$ denote the oriented scene surfaces,
$\mathcal{A}_L\subset\mathcal{A}$ the emitting surfaces, and
$\omega_{a\to b}=(b-a)/\|b-a\|_2$. For distinct surface points $x$ and $y$, the
solid angle subtended by an unoccluded differential area at $y$ satisfies
\[
\mathrm{d}\omega_i
=\frac{|n_y\!\cdot\!\omega_{y\to x}|}
{\|x-y\|_2^2}\,\mathrm{d}A(y).
\]
For reflection and emission on the front side of each oriented surface, only
positive cosines contribute; a two-sided surface can be treated as two oriented
sides. Combining the Jacobian, the receiver cosine, and binary visibility
$V_{\mathcal{G}}(x,y)$ gives~\citep{pbrt4}
\begin{equation}
\label{eq:geometry-kernel}
\begin{aligned}
G_{xy}^{\mathcal{G}}
&=\frac{V_{\mathcal{G}}(x,y)
(n_x\!\cdot\!\omega_{x\to y})_+
(n_y\!\cdot\!\omega_{y\to x})_+}
{\|x-y\|_2^2},\\
K_{xy}^{\mathcal{G},\mathcal{M}}(\omega_o)
&=f_r^{\mathcal{M}}(x,\omega_{x\to y},\omega_o)G_{xy}^{\mathcal{G}}.
\end{aligned}
\end{equation}
Radiance conservation along an unobstructed segment identifies incident radiance at
$x$ with outgoing radiance at $y$. The general surface balance, including emission
at a vertex that may itself be a light source, is
\begin{equation}
\label{eq:surface-recursion}
\begin{aligned}
L_o^{\mathcal{G},\mathcal{M},\Lambda}(x,\omega_o)
={}&L_e^{\Lambda}(x,\omega_o)\\
&+\int_{\mathcal{A}}K_{xy}^{\mathcal{G},\mathcal{M}}
(\omega_o)
L_o^{\mathcal{G},\mathcal{M},\Lambda}
(y,\omega_{y\to x})\,\mathrm{d}A(y).
\end{aligned}
\end{equation}

At the visible, non-emissive receiver in Eq.~\eqref{eq:render},
$L_e^{\Lambda}(x,\omega_o)=0$.

\paragraph{Direct and indirect transport: Eq.~\eqref{eq:render}.}
Define reflected radiance at a surface point by
$L_r=L_o-L_e^{\Lambda}$. Equation~\eqref{eq:surface-recursion} gives
\[
L_r(y,\omega_{y\to x})
=\int_{\mathcal{A}}K_{yz}^{\mathcal{G},\mathcal{M}}(\omega_{y\to x})
L_o(z,\omega_{z\to y})\,\mathrm{d}A(z).
\]
Here and below, omitted superscripts on $L_o$ and $L_r$ refer to the original
scene $(\mathcal{G},\mathcal{M},\Lambda)$. Substituting
$L_o(y,\omega)=L_e^{\Lambda}(y,\omega)+L_r(y,\omega)$ and using
$L_e^{\Lambda}=0$ outside $\mathcal{A}_L$ yields, at the non-emissive receiver,
\begin{equation}
\label{eq:direct-indirect-expansion}
\begin{aligned}
L_o(x,\omega_o)&=L_o^{\mathrm{dir}}(x,\omega_o)+L_o^{\mathrm{ind}}(x,\omega_o),\\
L_o^{\mathrm{dir}}(x,\omega_o)
&=\int_{\mathcal{A}_L}K_{xy}^{\mathcal{G},\mathcal{M}}(\omega_o)
L_e^{\Lambda}(y,\omega_{y\to x})\,\mathrm{d}A(y),\\
L_o^{\mathrm{ind}}(x,\omega_o)
&=\int_{\mathcal{A}}K_{xy}^{\mathcal{G},\mathcal{M}}(\omega_o)
L_r(y,\omega_{y\to x})\,\mathrm{d}A(y)\\
&=\int_{\mathcal{A}}K_{xy}^{\mathcal{G},\mathcal{M}}(\omega_o)
\left[\int_{\mathcal{A}}K_{yz}^{\mathcal{G},\mathcal{M}}
(\omega_{y\to x})
L_o(z,\omega_{z\to y})\,\mathrm{d}A(z)\right]
\mathrm{d}A(y).
\end{aligned}
\end{equation}
The direct term connects an emitter to $x$ and contains only the receiver
BRDF. The indirect term also includes the BRDF at $y$; its inner $L_o$
contains all higher-order reflections. Repeated substitution exposes the BRDF
at each intermediate vertex~\citep{veach1997}.

\paragraph{Canonical receiver: Eq.~\eqref{eq:canonical-render}.}
To make the intermediate operation precise, we hold the original incident field
fixed and evaluate a virtual receiver with the unit-albedo Lambertian BRDF
$f_r^{\mathrm{can}}=1/\pi$. Replacing only the outer receiver factor in
Eq.~\eqref{eq:direct-indirect-expansion} gives
\begin{equation}
\label{eq:canonical-area}
\begin{aligned}
L_{o,\mathrm{can}}^{\mathcal{G},\mathcal{M},\Lambda}(x,\omega_o)
={}&\frac{1}{\pi}\int_{\mathcal{A}_L}
G_{xy}^{\mathcal{G}}L_e^{\Lambda}(y,\omega_{y\to x})\,\mathrm{d}A(y)\\
&+\frac{1}{\pi}\int_{\mathcal{A}}
G_{xy}^{\mathcal{G}}L_r^{\mathcal{G},\mathcal{M},\Lambda}
(y,\omega_{y\to x})\,\mathrm{d}A(y).
\end{aligned}
\end{equation}
This is the surface-area form of Eq.~\eqref{eq:canonical-render}. The second
integral retains the original reflected field and its material dependence.
The canonical receiver is evaluated locally under that field; replacing all
scene materials and recomputing global illumination would define a different
quantity.

\paragraph{Direct-only representation: Eq.~\eqref{eq:diffdir}.}
Discarding the second integral in Eq.~\eqref{eq:canonical-area}, using target
light source parameters $\Lambda_t$, and evaluating at
$x_p=x_{\mathcal{G},\mathcal{C}}(p)$ gives
\begin{equation}
\label{eq:diffdir-area}
\begin{aligned}
L_t(p)
&=\frac{1}{\pi}\int_{\mathcal{A}_L(\Lambda_t)}
L_e^{\Lambda_t}(y,\omega_{y\to x_p})
G_{x_p y}^{\mathcal{G}}\,\mathrm{d}A(y)\\
&=\frac{1}{\pi}\int_{\mathcal{H}_{x_p}}
L_{i,\mathrm{src}}^{\mathcal{G},\Lambda_t}(x_p,\omega_i)
(n_{x_p}\!\cdot\!\omega_i)_+\,\mathrm{d}\omega_i.
\end{aligned}
\end{equation}
The second line recovers Eq.~\eqref{eq:diffdir}. Source emission and the
geometric kernel are independent of the original BRDFs, while geometry retains
its effect through visibility, orientation, and distance. Under the opaque-surface
assumption, replacing non-emitting materials by the canonical material and
extracting diffuse-direct transport gives the same quantity: no intermediate
surface is shaded along a retained path. Indirect illumination is excluded.

\subsection{Relation to Comparison Representations}

The representations contrasted in Section~\ref{sec:rep} can be related to different
simplifications or factorizations of the rendering equation~\eqref{eq:render}.
\emph{Irradiance} integrates the incident field without an explicit receiver
BRDF factor,
\begin{equation}
\label{eq:irradiance}
E^{\mathcal{G},\mathcal{M},\Lambda}(x) = \int_{\mathcal{H}_x}
L_i^{\mathcal{G},\mathcal{M},\Lambda}(x,\omega)
(n_x\!\cdot\!\omega)_+\,\mathrm{d}\omega,
\end{equation}
so the canonical response in Eq.~\eqref{eq:canonical-render} is $E/\pi$.
The incident field can still contain material-dependent interreflections, including
color bleeding. Moreover, changing a receiver material can change that field
through later bounces; the absence of an explicit local BRDF does not imply
invariance to such changes. In contrast, the lumi map keeps only direct-source
irradiance, divided by $\pi$.

For a non-emissive Lambertian surface with albedo $\rho(x)>0$, linear outgoing
radiance satisfies $L_o(x)=\rho(x)E(x)/\pi$. Under this assumption, the intrinsic
\emph{shading} factor is $S(x)=L_o(x)/\rho(x)=E(x)/\pi$, with division performed
per channel for RGB. Applying this factorization to general images requires an
albedo estimate and is not exact in the presence of non-Lambertian reflectance
or nonlinear image encoding. The division itself does not remove indirect light;
if $L_o$ is restricted to a diffuse-direct component, the corresponding $S$
instead represents direct irradiance divided by $\pi$.

The simplified \rev{cosine-only} \emph{Phong} \rev{feature used here} keeps the cosine term,
\begin{equation}
L_{\text{Phong}}(x) = \sum_{\ell=1}^{N_L} a_\ell\,
\max\!\bigl(0,\;n_x\!\cdot\!\omega_\ell(x)\bigr),
\label{eq:phong}
\end{equation}
with $a_\ell$ a source-control coefficient and $\omega_\ell(x)$ the direction
toward the source. This expression omits visibility and inverse-square falloff,
and thus cannot encode cast-shadow support or distance-dependent attenuation.
This comparison concerns the cosine-only feature. Lumi retains these geometric
effects through Eqs.~\eqref{eq:diffdir-area}. In Figure~\ref{fig:representation}, the scalar
Phong response is linearly normalized over valid pixels and the sRGB panel uses a
fixed display exposure; neither display transform is used as a model input.

\FloatBarrier
\section{Synthetic Data Generation}
\label{app:data}

\begin{figure}[t]
  \centering
  \includegraphics[width=.95\linewidth]{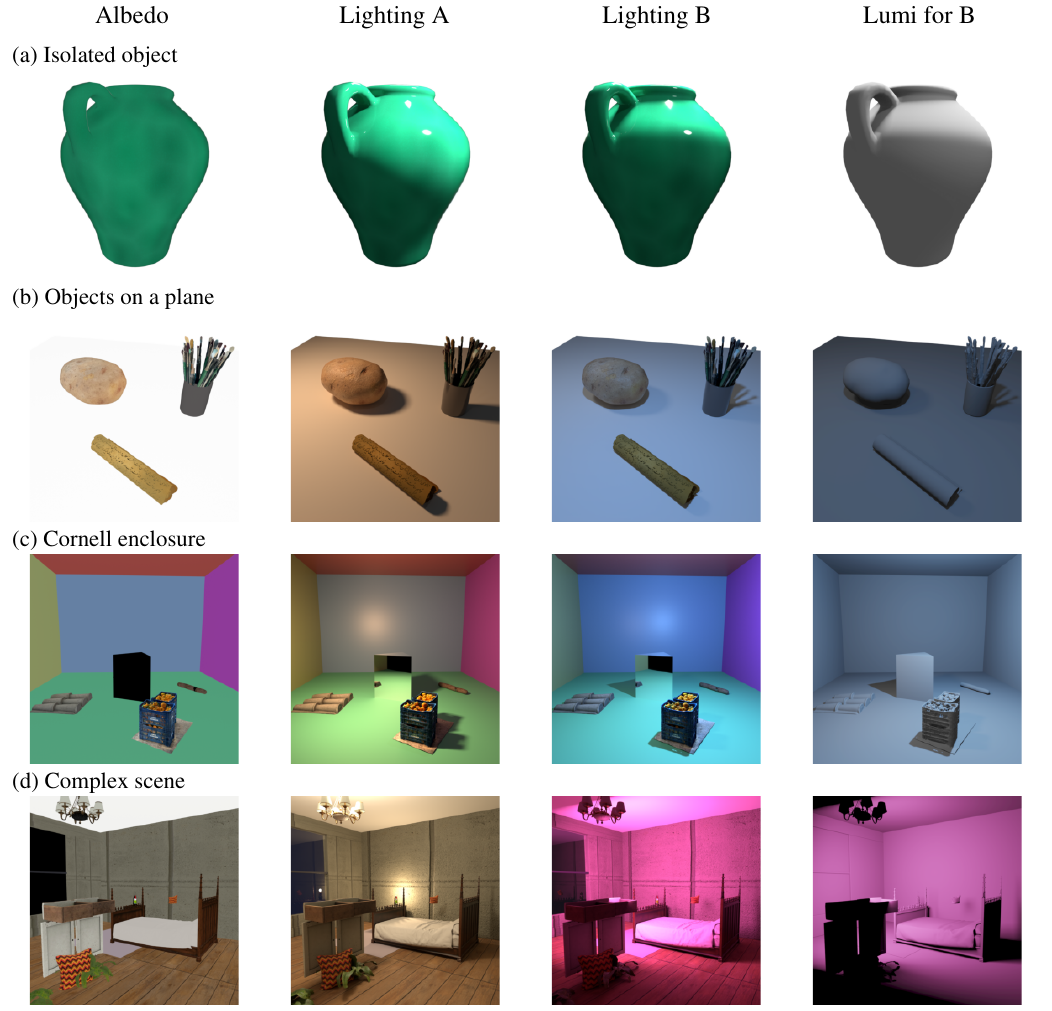}
  \caption{Examples from the four synthetic scene families. Each row shows albedo,
  two lighting conditions, and the lumi map for condition B. The object example
  varies the relative contributions of fixed light positions at fixed color and
  exposure. Lumi is displayed with $C_{\mathrm{display}}=C^{1/2.2}$.}
  \label{fig:appendix-data-examples}
\end{figure}

\paragraph{Scene composition.}
We construct four scene families: isolated objects, multiple objects on a ground
plane, objects in a Cornell-style enclosure, and authored indoor/outdoor scenes.
The first three use BlenderKit assets. Complex scenes combine procedural geometry
with PolyHaven materials, environment maps, and optional models.
Figure~\ref{fig:appendix-data-examples} illustrates the scene families and their
lighting variations, including self-shadowing, cast shadows, and occlusion.

\paragraph{Lighting bases and supervision.}
For object, plane, and Cornell scenes, we render RGB monochromatic light bases
at one to four positions for each camera, together with a separate environment
contribution. The hard subset uses four positions for object/plane scenes and two
or three for Cornell scenes. Beauty and lumi bases share the same geometry,
camera, and lighting configuration. During training, the loader combines both
sets of scene-linear bases with the same nonnegative RGB coefficients
(Appendix~\ref{app:model}). Complex scenes instead provide six authored lighting
states per group. Online mixing generates source/target pairs from the stored
bases, so training pairs are not counted as a fixed set of rendered image files.

Beauty retains the original materials and full light transport. Lumi is exported
from a separate pass with white diffuse materials (IOR 1, roughness 1, and
metallic 0), using Diffuse Direct without denoising. Albedo and available geometry
buffers are auxiliary outputs. Floating-point EXRs store scene-linear values;
exposure normalization and image encoding are applied by the training loader.
Table~\ref{tab:appendix-render} lists the rendering settings.

\begin{figure}[t]
  \centering
  \includegraphics[width=\linewidth]{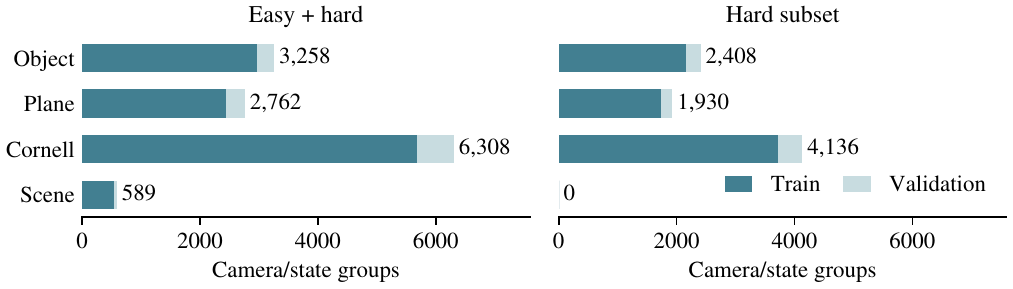}
  \caption{Composition of the synthetic data, counted in camera/state groups.
  Colors distinguish training and validation partitions. The hard collection is
  a subset of easy+hard; their counts should not be added.}
  \label{fig:appendix-data-counts}
\end{figure}

\paragraph{Data organization and split.}
A \emph{group} consists of a fixed model/scene and camera with its available
light bases or authored states. This unit differs from both a unique 3D asset and
an individual image. The easy+hard collection contains 12,917 groups
(11,626 training; 1,291 validation), including 589 complex-scene groups.
Its hard subset contains 8,474 groups (7,628 training; 846 validation), covering
object, plane, and Cornell scenes. Figure~\ref{fig:appendix-data-counts} shows the
composition. For each index separately, we sort the distinct relative model/scene
names, shuffle them with seed 0, and reserve a fraction of 0.1 for validation.
Note that the data counts reported here are consistent with those in the main text; 
any differences arise solely from different counting conventions. The training and test sets are disjoint.

The real-data reconstruction branch uses 38,640 pairs: 23,640 from MultiIllum
and 15,000 from ReNe. These are pair counts, separate from the synthetic groups.

\begin{table}[!hbp]
\centering
\caption{Synthetic rendering settings. Camera and light parameters that depend on
scene family describe the illustrated batches; the hard-subset position counts
apply to the corresponding indexed groups.}
\label{tab:appendix-render}
\small
\begin{tabular}{p{.25\linewidth}p{.7\linewidth}}
\toprule
Property & Setting \\
\midrule
Renderer / resolution & Blender Cycles; $1024\times1024$ \\
Sampling & Adaptive; maximum 512 samples, minimum 32, threshold 0.005 \\
Beauty/lumi storage & Scene-linear, 32-bit floating-point RGBA EXR \\
Canonical lumi & White material; IOR 1, roughness 1, metallic 0; Diffuse Direct only; no denoising \\
Object / plane optics & 35\,mm lens, $36\times24$\,mm sensor; sphere / upper-hemisphere camera sampling \\
Object / plane lights & POINT/AREA; hard groups: 4 positions; illustrated batch: 500\,W base power, area size 0.15--0.6 \\
Cornell optics / lights & 30/60\,mm focal settings; 2--3 light positions in hard groups; POINT/AREA, emitters camera-invisible \\
Complex scene example & 35\,mm camera; six named lighting states, including daylight, artificial light, and overcast conditions \\
\bottomrule
\end{tabular}
\end{table}

\section{Training Details}
\label{app:model}

\subsection{Lighting augmentation}
\begin{figure}[t]
  \centering
  \includegraphics[width=\linewidth]{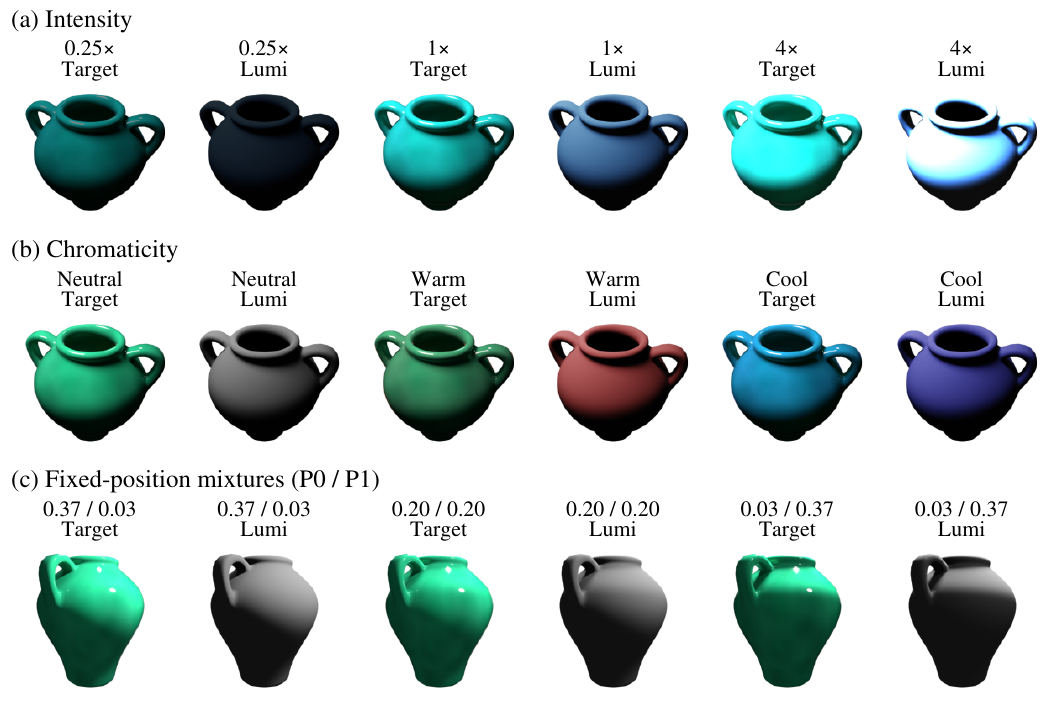}
  \caption{Lighting augmentation on the same object: (a) shared intensity scaling,
  (b) OLAT color changes, and (c) mixtures of two fixed light positions.
  Rows (b,c) keep exposure and environment illumination fixed; (c) uses another
  stored view and a fixed total position weight. Lumi is shown in its native
  control encoding.}
  \label{fig:appendix-augmentation}
\end{figure}

We vary the relative contributions of light positions, illumination intensity,
and chromaticity to expose the model to different lighting commands while
retaining the scene content. Figure~\ref{fig:appendix-augmentation} illustrates
these operations on the same object.

\paragraph{Combining light positions.}
For each OLAT camera group, source and target coefficients are sampled
independently. The beauty image is formed in linear RGB as
$B=\sum_{p,c}w_{pc}B_{pc}+B_{\mathrm{env}}$, with $w_{pc}\geq0$;
the same coefficients combine the lumi bases. To encourage changes in lighting direction, 
we select different preferred light positions for the source and target and scale the remaining light coefficients by 0.2.
Changing these weights varies the contribution of each fixed, pre-rendered
position. It does not move the light sources or subtract their images.

\paragraph{Intensity and chromaticity.}
At each light position, the sampler selects a color direction and intensity.
The ordinary branch samples a base coefficient in $[0.15,1]$ and chroma in
$[0,0.9]$. With probability 25\%, an extreme branch instead uses chroma in
$[0.85,1]$ and a channel ceiling of 1.8. Target mean luminance follows
$\mathcal{N}(0.18,0.06^2)$ clipped to $[0.05,0.45]$ in the ordinary branch,
and $\mathcal{U}[0.45,0.75]$ in the extreme branch. The exposure multiplier
is bounded to $[0.05,20]$. In the corrected paired loader, target beauty and
lumi receive the same multiplier before encoding.

\paragraph{Real-pair augmentation.}
The real-data branch estimates lumi from the target photograph using the same
model and detaches the estimate from the reconstruction gradient. It samples
$s\sim\operatorname{LogUniform}(0.25,4)$, scales the target photograph in
linear RGB, and converts it back to sRGB. The estimated lumi values are multiplied
directly by $s$ and clipped. The source photograph stays fixed. This provides
relative intensity commands without ground-truth lumi; the synthetic chromaticity
sampler is not applied to real photographs.

\subsection{Control encoding and sparsification}

\paragraph{Control sampling.}
\label{sec:method:control}
Within the synthetic relighting branch, half of the controls are freehand
strokes. The other half is divided equally among dense lumi and SEEDS superpixel
maps with three or nine luminance levels. Superpixel means provide local color,
while quantization coarsens magnitude. The stroke sampler uses 100 candidate
segments and 10 bins, a nominal budget of 5--20 strokes, lengths of 25--150 pixels,
and widths of 16--32 pixels. Color-axis anchors encourage coverage of chromatic
and bright/dark regions.

\begin{figure}[t]
  \centering
  \includegraphics[width=\linewidth]{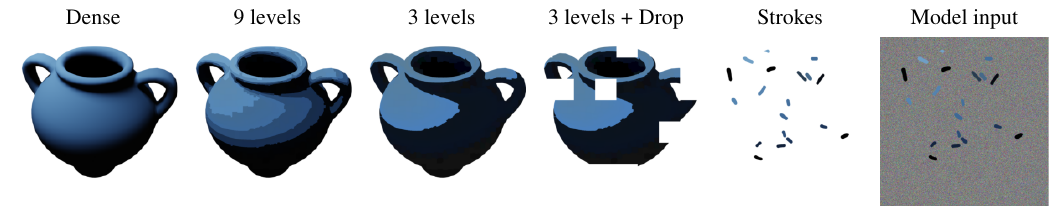}
  \caption{From dense lumi to sparse input. The alternatives include nine-level and
  three-level maps, a three-level map with patch dropping, and freehand strokes.
  The final panel noise-fills the plain stroke map.}
  \label{fig:sampling}
\end{figure}

Patch dropping is applied with probability 0.5. The map is divided into an
$8\times8$ grid; a drop probability is drawn uniformly from $[0,0.5]$, and each
cell is removed independently with that probability. This operation can also
remove parts of stroke maps. Missing pixels have zero alpha and are filled with
clipped Gaussian RGB noise (mean 127.5, standard deviation 64 on the 0--255 scale)
before model encoding. This distinguishes unspecified regions from specified
black illumination. Figure~\ref{fig:sampling} shows the control branches.

\subsection{Training configuration}
We adapt Qwen-Image-Edit-2511 with rank-64 LoRA.
Table~\ref{tab:appendix-training} summarizes the two-stage training schedule
specified in the main text. Stage 1 uses synthetic supervision, while Stage 2
adds reconstruction supervision from real image pairs. In both stages, 10\%
of synthetic items are assigned to control estimation and 90\% to relighting.
The real-pair reconstruction loss is applied every second Stage-2 update
with weight 1. Real images preserve their aspect ratio, with the longest
side capped at 512 pixels.

\begin{table}[!ht]
\centering
\caption{Training hyperparameters.}
\label{tab:appendix-training}
\small
\setlength{\tabcolsep}{4pt}
\renewcommand{\arraystretch}{1.15}
\begin{tabular}{@{}p{.30\linewidth}p{.335\linewidth}p{.315\linewidth}@{}}
\toprule
Setting & Stage 1 & Stage 2 \\
\midrule
LoRA rank & 64 & 64 \\
Initialization & Pretrained backbone & Corresponding Stage-1 LoRA \\
Training data & Synthetic & Synthetic + real pairs \\
Real training pairs & --- & 38,640 \\
Optimizer updates & 15,000 & 5,000 additional \\
Synthetic resolution & $512\times512$ & $1024\times1024$ \\
\midrule
Optimizer & AdamW & AdamW \\
Learning rate & $1\times10^{-4}$ & $5\times10^{-5}$ \\
LR schedule / warmup & Constant / none & Constant / none \\
Global synthetic batch size & 16 & 16 \\
Batch per GPU / accumulation & 1 / 1 & 1 / 1 \\
\midrule
Estimation / relighting & 0.1 / 0.9 & 0.1 / 0.9 \\
Real reconstruction loss & Disabled & Every 2 updates; weight 1 \\
\bottomrule
\end{tabular}
\par\smallskip
\begin{minipage}{\linewidth}
\footnotesize
\end{minipage}
\end{table}

\clearpage
\section{Additional Experiments}
\label{app:results}

\subsection{Evaluation protocol and conditioning}
\label{app:baseline-implementation}

\paragraph{Inputs.}
Table~\ref{tab:baseline-interfaces} lists each method's supported inputs.
RGB$\leftrightarrow$X and ScribbleLight receive dense lumi through their
irradiance and shading interfaces; Lume-Palette receives full-image lighting
maps. Our method receives sparse lumi strokes.

\begin{table}[!ht]\centering\scriptsize
\setlength{\tabcolsep}{3pt}
\begin{tabular}{@{}p{.14\linewidth}p{.25\linewidth}p{.25\linewidth}p{.30\linewidth}@{}}\toprule
Method & Source / auxiliary inputs & Prompt & Lighting condition\\\midrule
IC-Light & Source image & Selected lighting instruction & Text only; no direct control-map input.\\\addlinespace[3pt]
ScribbleLight & Source-derived albedo and normals & $p$ & Shading in the scribble/shading control slot.\\\addlinespace[3pt]
Gemini-Banana & Source as image 1 & $P_{\mathrm{B}}(p)$ & Dense lumi as image 2.\\\addlinespace[3pt]
Qwen-Image-Edit & Source as image 1 & $p$ + control-image directive & Dense lumi as image 2.\\\addlinespace[3pt]
GPT-Image-2 & Source as image 1 & $P_{\mathrm{G}}(p)$ & Dense lumi as image 2.\\\addlinespace[3pt]
RGB$\leftrightarrow$X & Source-derived albedo, normals, roughness, metallicness & $p$ for X-to-RGB & Irradiance in the irradiance slot.\\\addlinespace[3pt]
Lume-Palette & Source duplicated into four view slots; four canonical relit references & $p$ for casting; directional reference prompts & Lighting map replicated across four view slots.\\\addlinespace[3pt]
FLUX.2 Klein & Source as image 1 & $p$ + control-image directive & Dense lumi as image 2.\\\addlinespace[3pt]
\method{} & Source image & \texttt{relight the image} & Sparse RGBA lumi strokes.\\\bottomrule
\end{tabular}
\caption{Inputs, prompts, and lighting conditions for paired reconstruction. Prompt symbols are expanded below.
For hand-drawn in-the-wild tasks, image-conditioned methods instead receive the
lumi strokes, while IC-Light receives an annotated lighting instruction.}
\label{tab:baseline-interfaces}
\end{table}

\begin{figure}[!ht]
\centering
\includegraphics[width=\linewidth,height=0.35\textheight,keepaspectratio]{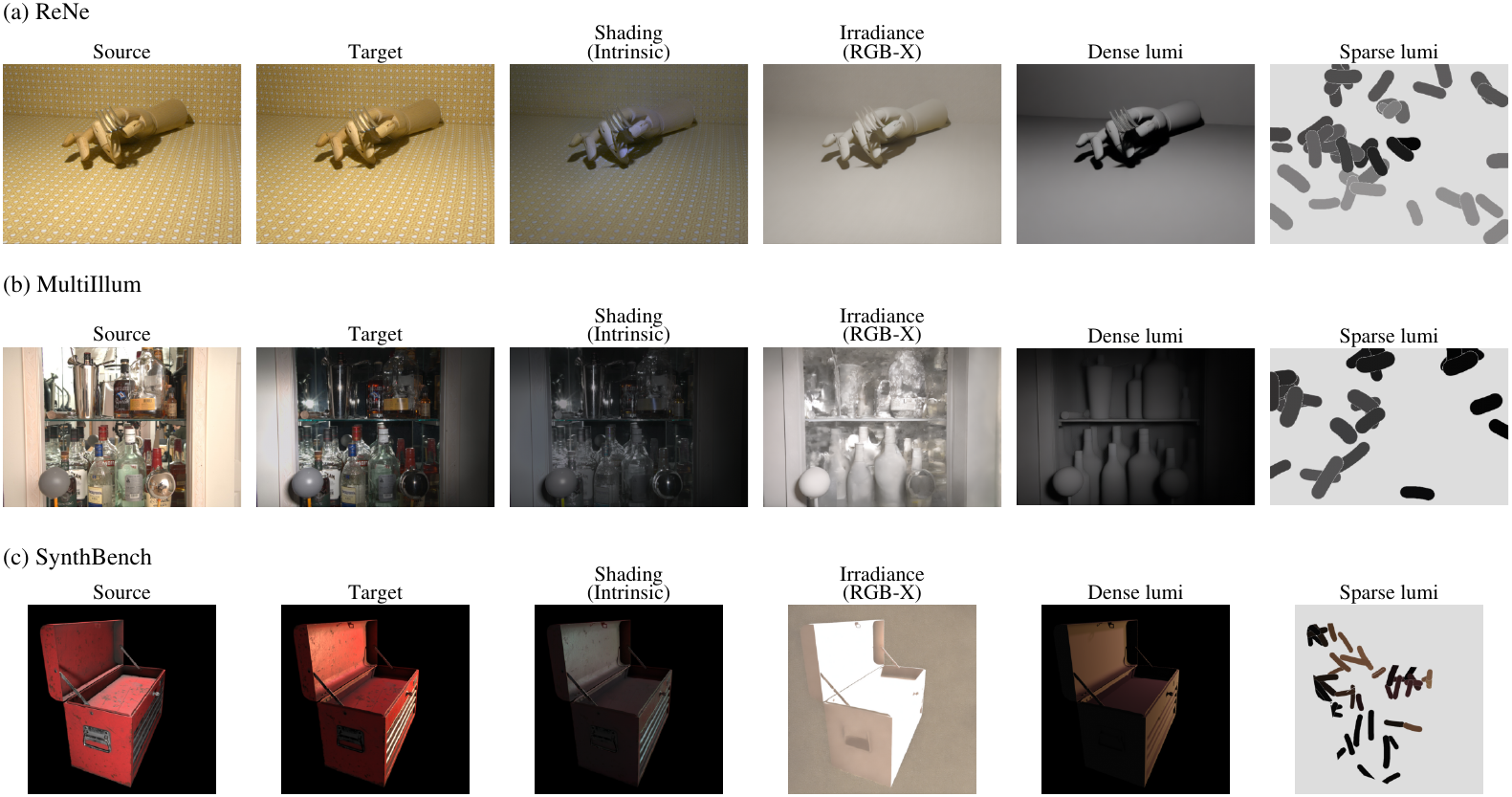}
\caption{Target lighting representations on ReNe, MultiIllum, and SynthBench.
Each row shows a source--target pair, shading estimated by Intrinsic~\citep{intrinsic},
irradiance estimated by RGB$\leftrightarrow$X~\citep{rgbx}, dense lumi, and sparse lumi
strokes. Both estimators use public pretrained weights. Gray denotes unmarked regions.}
\label{fig:app-evaluation-conditions}
\end{figure}

\clearpage

\FloatBarrier
\paragraph{Generation prompts.}
\label{app:baseline-prompts}
\begingroup
\sloppy
Here, $p$ denotes the sample's base text instruction, before applying a
method-specific prompt template:
\begin{itemize}
\setlength{\itemsep}{2pt}
\item \textbf{ReNe and MultiIllum.} Use the paired dataset's lighting-direction
annotation, in the form
\texttt{Relight the scene by moving the main light from \{source direction\}
to \{target direction\}.}
For example, one MultiIllum pair uses
\texttt{Relight the scene by moving the main light from the front-left at
horizontal eye-level to the back-right at horizontal eye-level.}
\item \textbf{SynthBench and in-the-wild inputs.} Use
\texttt{relight the image}. IC-Light uses the separate text selection below.
\end{itemize}
In the templates below, \texttt{\{instruction\}} is replaced by this text.
$P_{\mathrm{B}}(p)$ and $P_{\mathrm{G}}(p)$ denote the resulting full prompts
for Gemini-Banana and GPT-Image-2. Our method always uses
\texttt{relight the image}.
\begin{description}
\setlength{\itemsep}{3pt}
\setlength{\parsep}{0pt}
\item[IC-Light.]
Use an emitter-derived lighting description when available; otherwise prefer
the VLM lighting annotation, then $p$. For in-the-wild inputs, use the
annotated lighting instruction. Append \texttt{best quality}; the negative prompt is
\texttt{lowres, bad anatomy, bad hands, cropped, worst quality}.

\item[ScribbleLight.]
Use $p$.

\item[Gemini-Banana.]
Use $P_{\mathrm{B}}(p)$:
\begin{quote}\small\ttfamily\raggedright
You are an image relighting model. Given a source photo and a target lighting control map, regenerate the source under the target lighting.\par
Source image: [source image]\par
Target lighting control: [control image]\par
Re-light according to the light distribution in image 2.\par
Instruction: \{instruction\}\par
Return only the final relit RGB image.
\end{quote}

\item[Qwen-Image-Edit and FLUX.2 Klein.]
Append \texttt{Re-light according to the light distribution in image 2.}
to $p$. Qwen's negative prompt is a single space.

\item[GPT-Image-2.]
Use $P_{\mathrm{G}}(p)$:
\begin{quote}\small\ttfamily\raggedright
Edit Image 1 only. Image 1 is the source photograph whose subject, identity, geometry, materials, text, camera, crop, and background must remain unchanged.\par\smallskip
Image 2 is a sparse RGBA lumi map: colored strokes specify the desired light color, intensity, and spatial location; transparent pixels mean no lighting instruction.\par\smallskip
Relight Image 1 according to Image 2, converting the sparse strokes into coherent illumination, shading, highlights, and cast shadows. Do not draw or copy the strokes, checkerboard, transparency pattern, or control map into the result. Do not output a collage or Image 2.\par\smallskip
Task instruction: \{instruction\}. Return one final relit RGB photograph.
\end{quote}

\item[RGB$\leftrightarrow$X.]
The RGB-to-X prompts are \texttt{Albedo (diffuse basecolor)},
\texttt{Camera-space Normal}, \texttt{Roughness}, and \texttt{Metallicness}.
X-to-RGB uses $p$.

\item[Lume-Palette.]
Generate references with \texttt{light from the left},
\texttt{light from the middle}, \texttt{light from the right}, and
\texttt{light from the top}. Casting uses $p$ and an empty negative prompt.

\item[\method{}.]
Use \texttt{relight the image}.
\end{description}
\endgroup

\clearpage
\paragraph{In-the-wild evaluation.}
\label{app:itw-elo}
Given the source, control, and region locations and colors, GPT-5.2 compares
anonymous outputs. It observes each candidate independently, prioritizing
\rev{content preservation}, then spatial, \rev{chromatic}, and intensity response. Physical evidence
breaks response ties. Comparisons count only when both display orders agree
on the winner and decision level. Elo ratings are reported in
Table~\ref{tab:itw-elo}.

Excerpts from the observer and comparison prompts:
\begin{quote}\small\ttfamily\raggedright
You are an anonymous, single-candidate observer of scrib-controlled relighting.\par
Record task-grounded visual facts before ANY comparison with other candidates.
\end{quote}
\begin{quote}\small\ttfamily\raggedright
List every frozen region ID exactly once. Compare spatial, hue and strength
separately.\par\smallskip
\end{quote}

\paragraph{Elo aggregation.}
Ratings start at $R=1000$. \rev{We use 1,917 accepted pairwise comparisons.} For an accepted comparison of $A$ and $B$,
\[
E_A=\frac{1}{1+10^{(R_B-R_A)/400}},\qquad
R_A'=R_A+32(S_A-E_A),\quad R_B'=R_B-32(S_A-E_A),
\]
where $S_A\in\{0,\tfrac12,1\}$ denotes a loss, tie, or win.
Matches are sorted by identifier, shuffled with random seed 42, and processed
once.

\FloatBarrier
\subsection{Additional quantitative comparisons}
\label{app:dense-baselines}
Table~\ref{tab:dense-baselines} reports additional baseline comparisons using
the inputs in Table~\ref{tab:baseline-interfaces}.

\begin{table}[htbp]\centering\scriptsize
\setlength{\tabcolsep}{2.5pt}
\resizebox{\linewidth}{!}{\begin{tabular}{lrrrrrrrrrrrr}\toprule
& \multicolumn{4}{c}{ReNe} & \multicolumn{4}{c}{MultiIllum} & \multicolumn{4}{c}{SynthBench} \\
\cmidrule(lr){2-5}\cmidrule(lr){6-9}\cmidrule(lr){10-13}
Method & RMSE$\downarrow$ & PSNR$\uparrow$ & SSIM$\uparrow$ & LPIPS$\downarrow$ & RMSE$\downarrow$ & PSNR$\uparrow$ & SSIM$\uparrow$ & LPIPS$\downarrow$ & RMSE$\downarrow$ & PSNR$\uparrow$ & SSIM$\uparrow$ & LPIPS$\downarrow$ \\ \midrule
RGB$\leftrightarrow$X & 33.28 & 17.75 & 0.4395 & 0.4115 & 65.14 & 12.17 & 0.4888 & 0.5119 & 90.04 & 9.44 & 0.1547 & 0.8667 \\
Lume-Palette & 24.58 & 20.50 & 0.6662 & 0.3257 & 44.39 & 16.14 & 0.7354 & 0.3570 & 53.12 & 14.85 & 0.6504 & 0.2735 \\
FLUX.2 Klein & 64.20 & 12.20 & 0.1420 & 0.4079 & 74.53 & 11.28 & 0.4218 & 0.4964 & 63.09 & 12.72 & 0.5991 & 0.3621 \\
\method{} & \textbf{20.03} & \textbf{22.38} & \textbf{0.7542} & \textbf{0.1740} & \textbf{33.49} & \textbf{18.21} & \textbf{0.7659} & \textbf{0.2275} & \textbf{24.89} & \textbf{21.12} & \textbf{0.8637} & \textbf{0.1166} \\
\bottomrule\end{tabular}}
\caption{Additional paired reconstruction results on ReNe, MultiIllum, and
SynthBench.}
\label{tab:dense-baselines}\end{table}

\end{document}